\documentclass[10pt,twocolumn,letterpaper]{article}

\usepackage{wacv}
\usepackage{times}
\usepackage{epsfig}
\usepackage{graphicx}
\usepackage{amsmath}
\usepackage{amssymb}
\usepackage{booktabs}
\usepackage{makecell}
\usepackage[table]{xcolor}
\usepackage{bbding}
\usepackage{xcolor}
\usepackage{color, colortbl}
\definecolor{citecolor}{HTML}{2980b9}
\definecolor{linkcolor}{HTML}{c0392b}
\usepackage[accsupp]{axessibility} 

\wacvalgorithmstrack 

\def\wacvPaperID{1533} 

\wacvfinalcopy 

\ifwacvfinal
\usepackage[hidelinks,pagebackref=true,breaklinks=true,colorlinks,bookmarks=false,citecolor=citecolor,letterpaper=true,linkcolor=linkcolor]{hyperref}
\else
\usepackage[hidelinks,pagebackref=true,breaklinks=true,colorlinks,bookmarks=false,citecolor=citecolor,letterpaper=true,linkcolor=linkcolor]{hyperref}
\fi

\begin{document}

\title{Nearest Neighbors Meet Deep Neural Networks for Point Cloud Analysis}

\author{Renrui Zhang$^{1,2}$,\ \ Liuhui Wang$^{1,2}$,\ \ Ziyu Guo$^{1}$,\ \ Jianbo Shi$^{2,3}$\vspace{0.2cm}\\
$^1$Peking University,\ \ $^2$Heisenberg Robotics,\ \ $^3$University of Pennsylvania\\
{\tt\small \{1700012927, 1900012932\}@pku.edu.cn,\ \ jshi@seas.upenn.edu}
}

\maketitle
\thispagestyle{empty}

\begin{abstract}
Performances on standard 3D point cloud benchmarks have plateaued, resulting in oversized models and complex network design to make a fractional improvement. We present an alternative to enhance existing deep neural networks without any redesigning or extra parameters, termed as \textbf{S}patial-\textbf{N}eighbor Adapter (\textbf{SN-Adapter}). Building on any trained 3D network, we utilize its learned encoding capability to extract features of the training dataset and summarize them as prototypical spatial knowledge. For a test point cloud, the SN-Adapter retrieves $k$ nearest neighbors ($k$-NN) from the pre-constructed spatial prototypes and linearly interpolates the $k$-NN prediction with that of the original 3D network.   By providing complementary characteristics, the proposed SN-Adapter serves as a plug-and-play module to economically improve performance in a non-parametric manner. More importantly, our SN-Adapter can be effectively generalized to various 3D tasks, including shape classification, part segmentation, and 3D object detection, demonstrating its superiority and robustness. We hope our approach could show a new perspective for point cloud analysis and facilitate future research.
\end{abstract}

\section{Introduction}

3D vision has wide usage in robotics and AI. Many methods have been proposed to tackle 3D tasks, including object recognition~\cite{qi2017pointnet,qi2017pointnet++,simpleview,pointmlp,curvenet,zhang2022pointclip} and scene-level understanding~\cite{aldoma2012tutorial,verdoja2017fast,chen2019deep,zheng2013beyond,3detr,votenet}. Existing 3D methods are built upon the learnable deep neural networks and benefit from their abilities to process the irregular point clouds. Starting from the concise PointNet~\cite{qi2017pointnet}, later researches upgrade it with hierarchical architectures~\cite{qi2017pointnet++,pointmlp}, point-based convolutions~\cite{li2018pointcnn,thomas2019kpconv,xu2021paconv}, attention mechanisms~\cite{guo2021pct}, and so on~\cite{dgcnn,curvenet}. 

Recent works focused on inserting complicated modules or excessively increasing network parameters to boost benchmark scores. This trend has not only harmed the efficiency during training and inference, but gradually saturated the benchmarks. As examples for shape classification on ModelNet40~\cite{modelnet40}, CurveNet~\cite{curvenet} delicately explores a set of spatial curves for aggregating local geometry, which leads to 10$\times$ slower training and 20$\times$ slower inference compared to PointNet++~\cite{qi2017pointnet++}. PointMLP~\cite{pointmlp} brings +11.9M parameters for only +0.5\% accuracy boost, which increases 19$\times$ more model scales than its elite version~\cite{pointmlp}. Therefore, we ask the question: \textit{could we boost the performance of existing 3D networks at the least cost, even without additional parameters or re-training?}

\begin{figure*}[t!]
  \centering
    \includegraphics[width=1\textwidth]{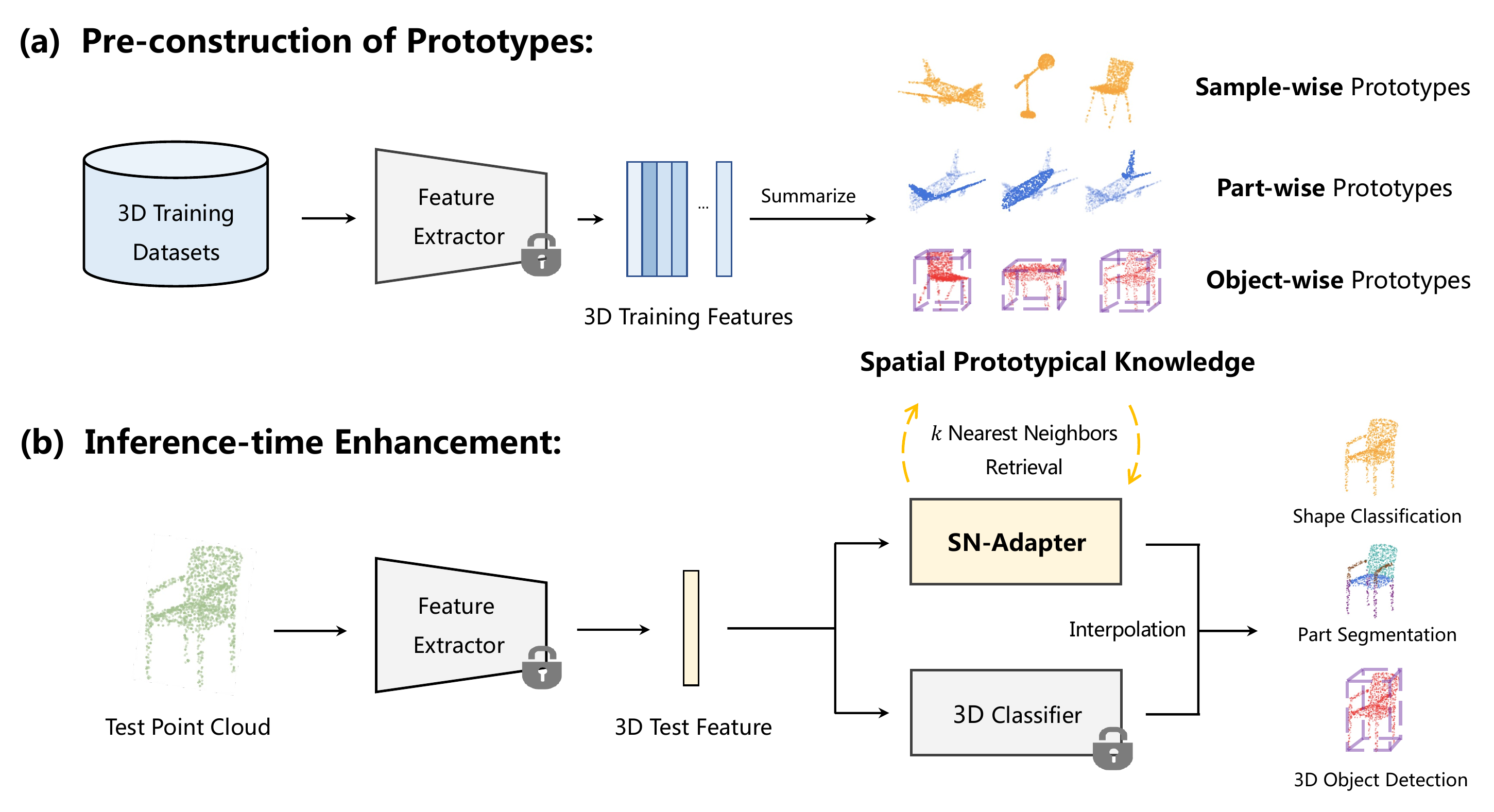}
   \caption{\textbf{The Pipeline of SN-Adapter by Two Steps: (a) and (b).} We divide the already trained deep neural network into the feature extractor and the 3D classifier, whose weights are frozen without fine-tuning.  In \textbf{a)}, we extract the 3D training features and construct the spatial prototypical knowledge. In \textbf{b)}, we introduce SN-Adapter to conduct $k$-NN retrieval from the task-specific prototypes for non-parametric enhancement.}
    \label{fig1}
\end{figure*}

We develop a non-parametric adapter module by retrieving 3D prototypical knowledge from the spatial neighbors, named \textbf{SN-Adapter}. It refers to the idea of $k$ Nearest Neighbors algorithm ($k$-NN) and can directly enhance the existing trained 3D deep neural networks without extra re-training. As shown in Figure~\ref{fig1}, our SN-Adapter is implemented in two steps: pre-construction of 3D prototypical knowledge and inference-time enhancement by interpolation. Specifically, we theoretically split a trained network into two parts. The first is the feature extractor that encodes an input raw point cloud into high-dimensional representations. The second, usually the last linear layer of the network, is named the 3D classifier, which categorizes the encoded vectors with classification logits. Using a trained extractor, we first obtain all the high-dimensional features of point clouds from the training dataset. For different 3D tasks, we summarize the features as various forms of prototypical spatial knowledge, e.g., sample-wise, part-wise, and object-wise prototypes in the top-right of Figure~\ref{fig1}. During inference, the SN-Adapter is appended to the feature extractor and utilizes $k$-NN to retrieve 3D knowledge from the pre-constructed prototypes. Finally, we linearly interpolate the classification logits concurrently produced from the SN-Adapter and the trained 3D classifier, by which the original 3D network can be improved with marginal extra costs.

Through experimental analysis, we observe the enhancement is resulted from the complementary characteristics between the trained 3D classifier and our SN-Adapter: the former is learned to fit the training set but the latter reveals the feature-level similarities among 3D prototypes. By extensive experiments, our SN-Adapter is verified to widely improve the performance of existing methods on different 3D tasks, such as +1.34\% classification accuracy on ModelNet40~\cite{modelnet40}, +0.17\% segmentation mIoU on ShapeNetPart~\cite{shapenetpart}, and +7.34\% detection AR on ScanNetV2~\cite{ScanNetV2}.



Our main contributions are summarized as follows:
\begin{enumerate}
    \item We propose SN-Adapter, a plug-and-play module to assist 3D deep neural networks via $k$-NN for better point cloud analysis.
    \item By retrieving knowledge from the pre-constructed spatial prototypes, SN-Adapter efficiently improves the already trained models without any parameters or re-training.
    \item We conduct complete experiments on various 3D benchmarks to demonstrate the effectiveness and robustness of our approach.
\end{enumerate}

\section{Related Work}
\paragraph{Deep Learning for 3D Point Clouds.}
Point cloud based shape classification of synthetic data~\cite{modelnet40} and real-world data~\cite{scanobjectnn} have been widely studied by PointNet~\cite{qi2017pointnet}, PointNet++~\cite{qi2017pointnet++} and so on~\cite{curvenet,dgcnn,pointmlp,li2018pointcnn,simpleview,zhang2021dspoint,zhang2022point,zhang2022learning}. Part segmentation~\cite{shapenetpart} and scene segmentation~\cite{ScanNetV2,sun_rgb} ask for the per-point classification, the methods~\cite{3detr,votenet,verdoja2017fast,chen2019deep,zheng2013beyond} of which normally extend feature decoders upon the classification networks to densely propagate the extracted features. 3D object detection has wide usages in \eg, autonomous driving~\cite{chen2017multi,navarro2010pedestrian,kidono2011pedestrian,huang2022tig} and robotics~\cite{rusu2009close,correll2016analysis,mousavian20196}.  Our SN-Adapter can be  generalized to all 3D tasks, including shape classification, part segmentation, and 3D object detection, demonstrating our robustness for point cloud analysis.

\paragraph{Feature Adapters in Computer Vision.}
The feature adapter is a light-weight module to efficiently adapt large-scale pre-trained models for downstream tasks. Motivated by the adapter in NLP~\cite{houlsby2019parameter}, CLIP-Adapter~\cite{gao2021clip}, Tip-Adapter~\cite{zhang2021tip}, and CoMo~\cite{zhang2022collaboration} introduce visual adapters using CLIP for few-shot image classification: freeze the pre-trained parameters of CLIP and only fine-tune the adapters of two-layer MLP. Follow-up works have successfully applied adapters to tasks such as 3D open-world learning~\cite{zhang2021pointclip,zhu2022pointclip}, image captioning~\cite{sung2022vl}, object detection~\cite{du2022learning}, semantic segmentation~\cite{rao2021denseclip}, and video analysis~\cite{wang2021actionclip}. Compared to previous works, our SN-Adapter is efficient, non-parametric, and aims at the tasks for 3D point clouds. We leverage the idea of $k$ nearest neighbors to enhance the trained 3D networks without re-training.

\paragraph{Nearest-Neighbor Algorithm.}
Nearest-Neighbor Algorithm memorizes the training data and predicts labels based on the $k$ nearest training samples ($k$-NN). Comparing with neural networks $k$-NN is still favored for its simplicity and efficiency. Models based on nearest-neighbor retrieval are able to provide strong baselines for many tasks, such as image captioning~\cite{devlin2015language, devlin2015exploring}, image restoration~\cite{plotz2018neural}, few-shot learning~\cite{wang2019simpleshot}, and representation learning~\cite{caron2021emerging, wallace2020extending}. Besides computer vision, Nearest-Neighbor Algorithm also plays an important role for some language tasks, \eg, language modeling~\cite{grave2017unbounded, khandelwal2019generalization} and machine translation~\cite{khandelwal2020nearest, tu2018learning}. Different from the above domains, for the first time, we explore how to augment existing deep neural networks with Nearest-Neighbor Algorithm for 3D point cloud analysis and propose an SN-Adapter with spatial prototypical knowledge retrieval.

\section{Method}
\label{s3}

In this section, we respectively illustrate how our proposed Spatial-Neighbor Adapter (SN-Adapter) benefits the three 3D tasks: shape classification, part segmentation, and 3D object detection.

\subsection{Shape Classification}
\label{s3.1}
\paragraph{Task Description.}
Given a trained 3D network for classification, we theoretically divide it into two parts: the feature extractor $\Phi(\cdot)$ and the 3D classifier $\Theta(\cdot)$. The feature extractor takes as input a raw point cloud $\{p_i\}_{i=1}^N$ of $N$ points and outputs its $C$-dimensional global feature $f \in \mathbb{R}^{C}$. The 3D classifier then maps $f$ into classification logits of $K$ categories, $l^{cls} \in \mathbb{R}^{K}$, which denote the predicted probability for each category. We formulate them as
\begin{align}
    l^{cls} = \Theta(f);\ \ \ f = \Phi(\{p_i\}_{i=1}^N).
\end{align}
Normally, $\Phi(\cdot)$ is invariant to the permutation of points with a pooling operation to capture the global characters, and $\Theta(\cdot)$ corresponds to the last linear projection layer of the network.

\paragraph{Sample-wise Spatial Prototypes.}
For shape classification, we construct the sample-wise spatial prototypes to retrieve 3D knowledge for each test point cloud. First, we utilize the trained feature extractor $\Phi(\cdot)$ to obtain the global features of all $M$ samples from the training set, denoted as $F^{cls} \in \mathbb{R}^{M \times C}$. As each training sample is only represented by a single global vector, we are affordable to store all $M$ features $F^{cls}$ as the spatial prototypes for reserving complete prior 3D knowledge, denoted as $Proto^{cls} \in \mathbb{R}^{M \times C}$. To further explore the spatial distributions of different point clouds, we also obtain a global positional vector for each training sample by averaging the 3D positional encodings~\cite{transformer} of all input points, which are directly added to $Proto^{cls}$. Then during inference, we extract the global feature $f$ of the test point cloud, and linearly interpolate the two classification logits predicted by the 3D classifier and our SN-Adapter, formulated as
\begin{align}
    l^{cls} = \Theta(f) + \gamma\text{SN-Adapter}(f,\ Proto^{cls}),
\end{align}
where $\gamma$ denote the relative weights between two logits.

\paragraph{SN-Adapter.}
Analogous to all 3D tasks, SN-Adapter conducts $k$-NN algorithm to aggregate $k$-nearest spatial knowledge and adopts Euclidean distance as the distance metric between $f$ and $Proto^{cls}$.
We represent the retrieved $k$-nearest prototypes as $\mathcal{N}$ and the category set as $\mathcal{C}$. Then, the predicted probability of category $c \in \mathcal{C}$ in the logits is calculated as
\begin{align}
    Prob(c|f) = \frac{\sum_{pt \in \mathcal{N}_c}1/d(f, pt)}{\sum_{c \in C}\sum_{pt \in \mathcal{N}_c}1/d(f, pt)},
\end{align}
where $\mathcal{N}_c$ denotes the retrieved prototypes of the $c$ category, and $d(f, pt)$ denotes the distance between test point cloud's feature $f$ and the prototype $pt$.

\subsection{Part Segmentation}
\label{s3.2}
\paragraph{Task Description.}
Part segmentation task requires the network to classify each point in the input point cloud. The $\Phi(\cdot)$ is developed as an encoder-decoder architecture and outputs the extracted features $\{f_i\}_{i=1}^N$ for all the $N$ points. We formulate this as
\begin{align}
    \{l^{seg}_i\}_{i=1}^N = \{\Theta(f_i)\}_{i=1}^N;\ \ \ \{f_i\}_{i=1}^N = \Phi(\{p_i\}_{i=1}^N),
\end{align}
where $l^{seg}_i \in \mathbb{R}^{K}$ denotes the classification logits of the $i$-th point. Here, $\Theta(\cdot)$ is shared for every point and maps the point feature into logits of $K$ part categories.

\paragraph{Part-wise Spatial Prototypes.}
We construct the part-wise spatial prototypes to retrieve 3D knowledge for every single point of the test point cloud. Considering the classification logits are to be made for each point, we need to extract and memorize the features $F$ of all $N$ points from $M$ training samples as prototypical knowledge. However, it would be overloaded to store the $F^{seg} \in \mathbb{R}^{M \times N \times C}$, let alone the $k$-NN retrieval. Therefore, for each training sample, we propose to obtain its part-wise prototypical features by conducting average pooling on the points of the same part category, denoted as $\text{Part\_Pooling}(\cdot)$. For example, a point cloud of a chair is annotated as three parts: leg, seat, and back. Then, we only need to store three prototypical features for this training sample, whose dimension is $\mathbb{R}^{3 \times C}$. After the pre-construction, we acquire the spatial prototypical knowledge for part segmentation, $Proto^{seg}$, which is space-efficient and still in the same order as $Proto^{cls}$, formulated as 
\begin{align}
    Proto^{seg} = \text{Part\_Pooling}(F^{seg}) \in \mathbb{R}^{M \times P\times C},
\end{align}
where $P$ is the maximum part category number of an object in the dataset, which is no more than six in ShapeNetPart~\cite{shapenetpart}. During inference, after extracting the features $\{f_i\}_{i=1}^N$, we combine the two classification logits for each $N$ point of the test point cloud, formulated as
\begin{align}
\begin{split}
    \{l^{seg}_i\}_{i=1}^N &= \{\Theta(f_i) \\&+ \gamma\text{SN-Adapter}(f_i,\ Proto^{seg})\}_{i=1}^N.
\end{split}
\end{align}

\subsection{3D Object Detection}
\label{s3.3}
\paragraph{Task Description.}
Taking a scene-level point cloud as input, the 3D object detector learns to localize and classify the objects in the 3D space. The detector would first utilize the $\Phi(\cdot)$ to extract the scene-level 3D features and group the features for each object proposal, denoted as $\{f_i\}_{i=1}^O$, where $f_i \in \mathbb{R}^{C}$ and $O$ represents the proposed object number of the scene. Then, several parallel MLP-based heads are adopted to predict the category, 3D position and other attributes for each object proposal. We formulate the main process as
\begin{align}
\begin{split}
    \{l^{det}_i\}_{i=1}^O &= \{\Theta_{cls}(f_i)\}_{i=1}^O;\ \ \ \\ \{p^{det}_i\}_{i=1}^O &= \{\Theta_{pos}(f_i)\}_{i=1}^O;\ \ \ \\ \{f_i\}_{i=1}^O &= \Phi(\{p_i\}_{i=1}^N),
\end{split}
\end{align}
where $\Theta_{cls}(\cdot)$ and $\Theta_{pos}(\cdot)$ are responsible for predicting the classification logits $l^{det}_i \in \mathbb{R}^{K}$ and the 3D position $p^{det}_i \in \mathbb{R}^{3}$, which are shared for all object proposals. After this, the Non-Maximum Suppression (3D NMS) is applied to discard the duplicated predictions in the 3D space, which is significant to the final evaluation metric.

\paragraph{Object-wise Spatial Prototypes.}
We construct the object-wise spatial prototypes to retrieve 3D knowledge for each object proposal in the test point cloud. We first leverage the trained 3D detector to obtain the extracted object features and predicted 3D positions for all training samples, denoted as $F^{det}, P^{det} \in \mathbb{R}^{M \times O \times 3}$. On top of that, we adopt positional encodings~\cite{transformer} based on trigonometric functions to embed $P^{det}$ and add them onto $F^{det}$. This provides $F^{det}$ with sufficient 3D positional information of objects and facilitates the $k$-NN retrieval for SN-Adapter. We then calculate the spatial prototypical knowledge for 3D object detection as
\begin{align}
    Proto^{det} = F^{det} + \text{PE}(P^{det}) \in \mathbb{R}^{M \times O\times C},
\end{align}
where PE denotes the positional encodings function.

During inference, for each object proposal we acquire its predicted $f_{i}, l_{i}^{det}, p_{i}^{det}$ and aggregate them likewise via positional encodings.  The SN-Adapter retrieves spatial knowledge from nearest neighbors and enhances the classification logits predicted by $\Theta_{cls}$, formulated as
\begin{align}
    \{l^{det}_i\}_{i=1}^O &= \{\Theta_{cls}(f_i)\\ &+ \gamma\text{SN-Adapter}(f_i + \text{PE}(p_i^{det}),\ Proto^{det})\}_{i=1}^O.\nonumber
\end{align}
Our SN-Adapter is inserted before the 3D NMS operation, which could rectify some `false' classification made by $\Theta_{cls}$ and effectively avoid the removal of `true' bounding boxes.

\begin{figure}[t]
    \centering
    \includegraphics[width=0.4\textwidth]{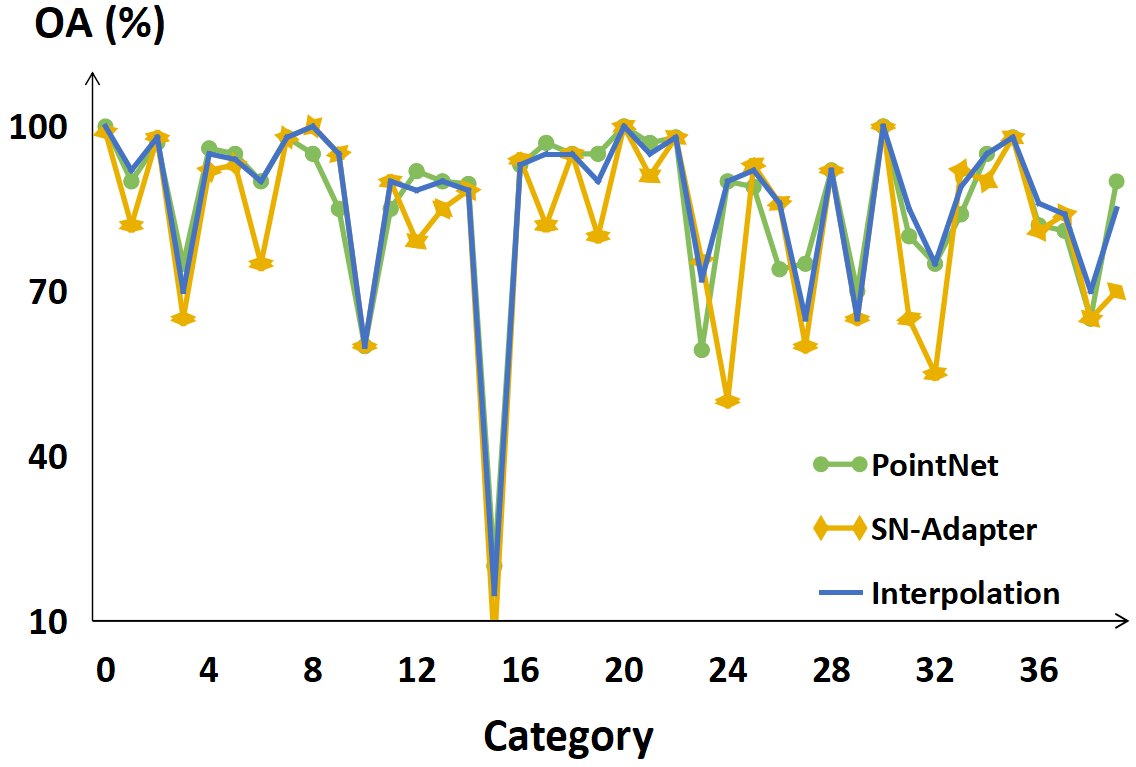}
    \vspace{0.1cm}
    \caption{\textbf{Comparison of individual PointNet, SN-Adapter, and the interpolated model for different categories.} We show the the overall accuracy (OA) of 40 categories on ModelNet40~\cite{modelnet40}.}
    \label{fig2}
\end{figure}

\begin{table}[t]
\small
\centering
\begin{tabular}{cccc}
	\toprule
		PointNet & SN-Adapter & Interpolation & Number\\
		\cmidrule(lr){1-1} \cmidrule(lr){2-2} \cmidrule(lr){3-3} \cmidrule(lr){4-4}
        \Checkmark & \XSolidBrush & \Checkmark & 72 \\
        \Checkmark & \XSolidBrush & \XSolidBrush & 36\\
        \XSolidBrush & \Checkmark & \Checkmark & 68 \\
        \XSolidBrush & \Checkmark & \XSolidBrush & 10\\
        \XSolidBrush & \XSolidBrush & \Checkmark & 1 \\
	  \bottomrule
	\end{tabular}
\vspace{0.2cm}
\caption{\textbf{Statistic of sample numbers where individual models produce different predictions.} \Checkmark and \XSolidBrush denote correct and wrong predictions, respectively.}
\label{t1}
\end{table}

\section{Analysis}

\subsection{Quantative Analysis} 

\begin{figure*}[t!]
  \centering
    \includegraphics[width=0.95\textwidth]{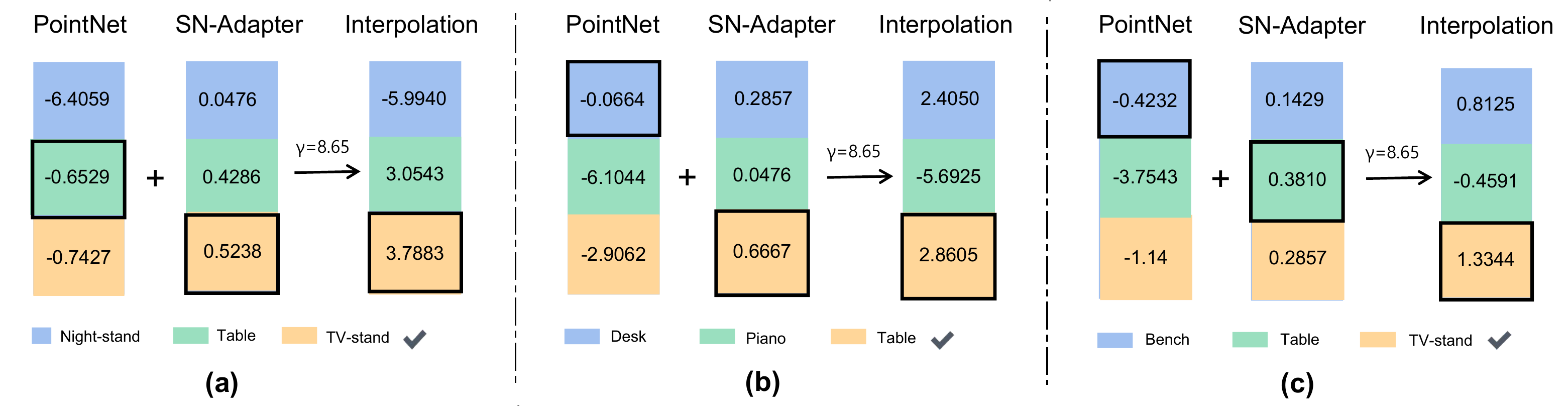}
    \vspace{0.1cm}
   \caption{\textbf{Classification logits of PointNet, SN-Adapter and the interpolated model}. We report the numerical results of logits before the softmax function and denote different categories with different colors. We highlight the category of the highest value in the logits with a box and the ground-truth category with a check mark.}
    \label{fig3}
\end{figure*}

Here, we take PointNet~\cite{qi2017pointnet} for shape classification on ModelNet40~\cite{modelnet40} as the example. First, we show the performance of individual PointNet's 3D classifier and SN-Adapter compared to the interpolated one in Figure~\ref{fig2}: the interpolated model achieves higher accuracy for most categories. Though SN-Adapter performs much worse than the learned 3D classifier on some categories, it could reversely enhance to achieve a better 3D classifier by interpolation. Specifically, we present the statistic for the interpolated prediction whose individual predictions of 3D classifier and SN-adapter are inconsistent. As shown in Table~\ref{t1}, when the original PointNet is wrong, but SN-Adapter is correct, our SN-Adapter can help rectify nearly 90\%, 68/(68+10), of the predictions. More surprisingly, we observe that, even if both PointNet and SN-Adapter are wrong, the interpolated one still obtains the correct results, demonstrating the implicit complementary knowledge between the learned 3D classifier and the spatial prototypes. 


To further illustrate the complementarity of SN-Adapter, we present the predicted classification logits before softmax function, for the cases where SN-Adapter corrects the false prediction of PointNet. As shown in Figure~\ref{fig3} (a), the PointNet's predicted values of `night-stand' and `table' are close, indicating it is difficult for PointNet to distinguish them. In contrast, the SN-Adapter could produce more discriminative values between the two categories and address the ambiguity of PointNet by an ensemble with a large weight $\gamma$. As for Figure~\ref{fig3} (b), when PointNet confidently predicts the wrong category, our SN-Adapter can put the final prediction back on track with the confidence score for the correct category. Figure~\ref{fig3} (c) shows that when both of their predictions are wrong, the interpolation of SN-Adapter can still contribute to the right answer.

\subsection{Qualitative Analysis}
Why does the $k$-NN retrieval work for point cloud analysis? For one, due to the difficulty of data acquisition, the 3D community lacks large-scale high-quality training datasets, and existing methods can only learn from limited samples.  In this situation, the representative 3D prototypes become much more significant since the construction of prototypes are not overly dependent on data distribution and can well represent the typical features of a category.  In contrast, the 3D classifiers of deep neural networks greatly suffer from long-tail distributions of training data. That is to say, when the 3D samples of some categories are insufficient during training, the learnable classifier would not form the prediction preference of those unusual categories and fail to recognize them for testing.  The $k$-NN upon spatial prototypes inherently overcomes such category imbalance via similarity-based retrieval, which hardly depends on the amount of training data.

\subsection{Theoretical Analysis}
We start from the perspective of learned embedding space to illustrate how SN-Adapter boosts the learned deep neural networks.
The $k$-NN algorithm of SN-Adapter is able to associate pre-constructed spatial prototypes in close proximity. These adjacent prototypes normally have the same ground-truth labels and share similar semantic knowledge. Spatially, the entire 3D space can be divided into many discrete spherical regions. We define a spherical region in the embedding space as $N_{\epsilon}(x) = \{x'||x'-x||_2 \leq \epsilon\}$, where $x$ denotes the spherical center and $\epsilon$ denotes its radius. The goal of SN-Adapter is based on the extracted feature of a test point cloud to retrieve clusters and then obtain representative knowledge from them. 

For better retrieval performance, each spherical center prefers to have sufficiently pure spherical regions. In other words, the information of the representative prototypes should be convincing enough, formulated as $\forall x'\in N_{\epsilon}(x), gt(x')=gt(x)$, where $gt(\cdot)$ denotes the ground-truth label. We then define $C(N_{\epsilon})$ and $P(N_{\epsilon})$ as the coverage and purity of all spherical regions. The optimal $C(N_{\epsilon})$ desires $\epsilon$ to be large enough to cover the entire space, while the purity requires a smaller $\epsilon$ to contain as few deviating prototypes as possible. Therefore, we need to consider the trade-off between both coverage and purity. Formally, we expect to obtain the specific $\epsilon$ that satisfies $\epsilon^* = max\{\epsilon: P(N_\epsilon)\geq\alpha\}$, where $\alpha$ serves as a threshold for $P(N_{\epsilon})$ and also the maximum function that helps increase $C(N_{\epsilon})$. In our experiments, we do not explicitly set the value $\alpha$, but leverage an appropriate number of $k$ nearest neighbors to implicitly obtain the optimal trade-off for better retrieving prototypical knowledge.

\section{Experiments}
\subsection{Shape Classification}
\paragraph{Settings}
We evaluate our SN-Adapter on two widely adopted datasets for shape classification: ModelNet40~\cite{modelnet40} and ScanObjectNN~\cite{scanobjectnn}. We select several representative methods and append SN-Adapter upon them: PointNet~\cite{qi2017pointnet}, PointNet++~\cite{qi2017pointnet++}, SpiderCNN~\cite{xu2018spidercnn}, DGCNN~\cite{dgcnn}, PCT~\cite{guo2021pct}, CurveNet~\cite{curvenet}, and PointMLP~\cite{pointmlp}. We set the last linear layer as $\Theta(\cdot)$ and all the precedent layers as $\Phi(\cdot)$. The overall accuracy (OA) and class-average accuracy (mAcc) are adopted as evaluation metrics. Note that, as our SN-Adapter requires no training time, we utilize simple loops to search for the best $k$ within minutes.

\begin{table}[t]
\vspace{0.1cm}
\small
\centering
	\begin{tabular}{lccc}
	\toprule
		\makecell*[c]{Method} &OA (\%) &mAcc (\%) &$k$\\
		\cmidrule(lr){1-1} \cmidrule(lr){2-2} \cmidrule(lr){3-3} \cmidrule(lr){4-4}
	    PointNet~\cite{qi2017pointnet}  &89.34 &85.79 &-\\\rowcolor{gray!12}\vspace{0.1cm}
	    \ \ \ \ + SN-Adapter &\textbf{90.68} &\textbf{86.47} &21 \\
	    PointNet++~\cite{qi2017pointnet++} &92.42 &89.22 &-\\\rowcolor{gray!12}\vspace{0.1cm}
	    \ \ \ \ + SN-Adapter &\textbf{93.48} &\textbf{90.00} &77 \\
	    DGCNN~\cite{dgcnn} &92.18 &89.10 &-\\\rowcolor{gray!12}\vspace{0.1cm}
	    \ \ \ \ + SN-Adapter &\textbf{92.99} &\textbf{89.70} &24 \\
	    PCT~\cite{guo2021pct} &93.27 &89.99 &-\\\rowcolor{gray!12}\vspace{0.1cm}
	   \ \ \ \ + SN-Adapter &\textbf{93.56} &\textbf{90.17} &110 \\
	    CurveNet~\cite{curvenet} &93.84 &91.14 &-\\\rowcolor{gray!12}
	    \ \ \ \ + SN-Adapter &\textbf{94.25} &\textbf{91.50} &2 \\
	  \bottomrule
	\end{tabular}
\vspace{0.2cm}
\caption{\textbf{Shape classification on ModelNet40~\cite{modelnet40}} dataset.}
\label{t2}
\end{table}
\begin{table}
\small
\centering
\begin{tabular}{lccc}
	\toprule
		\makecell*[c]{Method} &OA (\%) &mAcc (\%) &$k$\\
		\cmidrule(lr){1-1} \cmidrule(lr){2-2} \cmidrule(lr){3-3} \cmidrule(lr){4-4}
	    PointNet~\cite{qi2017pointnet}  &68.2 &63.4 &-\\\rowcolor{gray!12}\vspace{0.1cm}
	    \ \ \ \ + SN-Adapter &\textbf{70.1} &\textbf{64.2} &128 \\
	    SpiderCNN~\cite{xu2018spidercnn} &73.7 &69.8 &-\\\rowcolor{gray!12}\vspace{0.1cm}
	    \ \ \ \ + SN-Adapter &\textbf{74.4} &\textbf{70.5} &68 \\
	    PointNet++~\cite{qi2017pointnet++} &77.9 &75.4 &-\\\rowcolor{gray!12}\vspace{0.1cm}
	    \ \ \ \ + SN-Adapter &\textbf{79.2} &\textbf{76.2} &16 \\
	    DGCNN~\cite{dgcnn} &78.1 &73.6 &-\\\rowcolor{gray!12}\vspace{0.1cm}
	    \ \ \ \ + SN-Adapter &\textbf{78.9} &\textbf{74.0} &140 \\
	    PointMLP~\cite{pointmlp} &85.7 &84.0 &-\\\rowcolor{gray!12}
	    \ \ \ \ + SN-Adapter &\textbf{86.3} &\textbf{84.6} &5 \\
	  \bottomrule
\end{tabular}
\vspace{0.2cm}
\caption{\textbf{Shape classification on ScanObjectNN~\cite{scanobjectnn}} dataset.}
\label{t3}
\end{table}

\paragraph{Performance}
In Table~\ref{t2} and Table~\ref{t3}, we show the enhancement results of SN-Adapter on the two datasets, respectively. On ModelNet40~\cite{modelnet40} of synthetic data, PointNet++ is boosted by +1.06\% mAcc, which has surpassed the more complicated DGCNN by +1.30\%. On ScanObjectNN~\cite{scanobjectnn} of real-world data, SN-Adapter shows stronger complementary characteristics to the trained networks, which boosts PointNet by 1.9\% OA and PointNet++ by +1.3\% OA. For the state-of-the-art PointMLP, our SN-Adapter improves it by +0.6\% OA and +0.6\% mAcc. 

\subsection{Part Segmentation}
\paragraph{Settings}
For part segmentation, we test our SN-Adapter on ShapeNetPart~\cite{shapenetpart} dataset and select the four baseline models: DGCNN~\cite{dgcnn}, PointNet++~\cite{qi2017pointnet++}, PointMLP~\cite{pointmlp} and CurveNet~\cite{curvenet}. We follow other settings the same as the shape classification experiments and report the mean IoU
across all instances in the dataset, denoted as mIoU$_I$.

\vspace{-0.5cm}
\paragraph{Performance}
As the part segmentation benchmark has long been saturated, a slight improvement for mIoU$_I$ is also worth mentioning. In Table~\ref{t4}, we observe the biggest improvement of +0.17\% mIoU$_I$ is on PointMLP, compared to Curvenet's +0.11\% and DGCNN's +0.09\%. This indicates that the stronger feature encoder $\Phi(\cdot)$ contributes to better part-wise prototypes for the retrieval of SN-Adapter.

\begin{table}[t]
\vspace{0.1cm}
\small
\centering
\begin{tabular}{lcc}
	\toprule
		\makecell*[c]{Method} &mIoU$_I$ (\%) &$k$\\
		\cmidrule(lr){1-1} \cmidrule(lr){2-2} \cmidrule(lr){3-3}
	    DGCNN~\cite{dgcnn} &85.17  &-\\\rowcolor{gray!12}\vspace{0.1cm}
	    \ \ \ \ + SN-Adapter &\textbf{85.26}  &22 \\
	    PointNet++~\cite{qi2017pointnet++} &85.40 &-\\\rowcolor{gray!12}\vspace{0.1cm}
	    \ \ \ \ + SN-Adapter &\textbf{85.47}  &1 \\
	    PointMLP~\cite{pointmlp} &85.69  &-\\\rowcolor{gray!12}
	    \ \ \ \ + SN-Adapter  &\textbf{85.86} &1 \\
	    CurveNet~\cite{curvenet}  &86.58 &-\\\rowcolor{gray!12}\vspace{0.1cm}
	    \ \ \ \ + SN-Adapter &\textbf{86.69}  &64 \\
	  \bottomrule
	 \vspace{0.2cm}
\end{tabular}
\caption{\textbf{Part segmentation on ShapeNetPart~\cite{shapenetpart}}.}
\label{t4}
\end{table}
\begin{table}
\small
\centering
\begin{tabular}{lcc}
	\toprule
		\makecell*[c]{Method} &AP$_{25}$ (\%) &AR$_{25}$ (\%)\\
		\cmidrule(lr){1-1} \cmidrule(lr){2-2} \cmidrule(lr){3-3}
	    VoteNet~\cite{votenet} &57.84 &80.92 \\\rowcolor{gray!12}\vspace{0.1cm}
	    \ \ \ \ + SN-Adapter &\textbf{58.46} &\textbf{83.74}  \\
	    3DETR-m~\cite{3detr} &64.60 &77.22 \\\rowcolor{gray!12}
	    \ \ \ \ + SN-Adapter &\textbf{65.16} &\textbf{84.56} \\
	  \bottomrule
	  \vspace{0.2cm}
	\end{tabular}
\caption{\textbf{3D object detection on ScanNetV2~\cite{ScanNetV2}}.}
\label{t5}
\end{table}

\subsection{3D Object Detection}
\paragraph{Settings}
For 3D object detection on ScanNetV2~\cite{ScanNetV2}, we select VoteNet~\cite{votenet} and 3DETR-m~\cite{3detr} as the baseline models to test our SN-Adapter. We set the MLP-based classification head as $\Theta(\cdot)$ and the scene-level feature extractor as $\Phi(\cdot)$. The SN-Adapter is inserted after $\Theta(\cdot)$ and before the 3D NMS. We report the mean Average Precision (AP$_{25}$) and mean Average Precision (AR$_{25}$) at 0.25 IoU threshold. For time efficiency, the hyperparameter $k$ is simply set as 32 for the two detectors.

\vspace{-0.4cm}
\paragraph{Performance}
Table~\ref{t5} presents the enhanced detection performance by SN-Adapter. For AR$_{25}$, we significantly improve by +2.82\% on VoteNet and +7.34\% on 3DETR-m. This indicates that the spatial prototypical knowledge can effectively avoid the removal of false duplicated bounding boxes in 3D space. More specifically, some spatially neighboring boxes, which have incorrectly similar scores and should have been removed by 3D NMS, can be rectified and reserved as outputs. 

\subsection{Ablation Study}
\begin{figure}[t]
    \centering
    \includegraphics[height=1.8in]{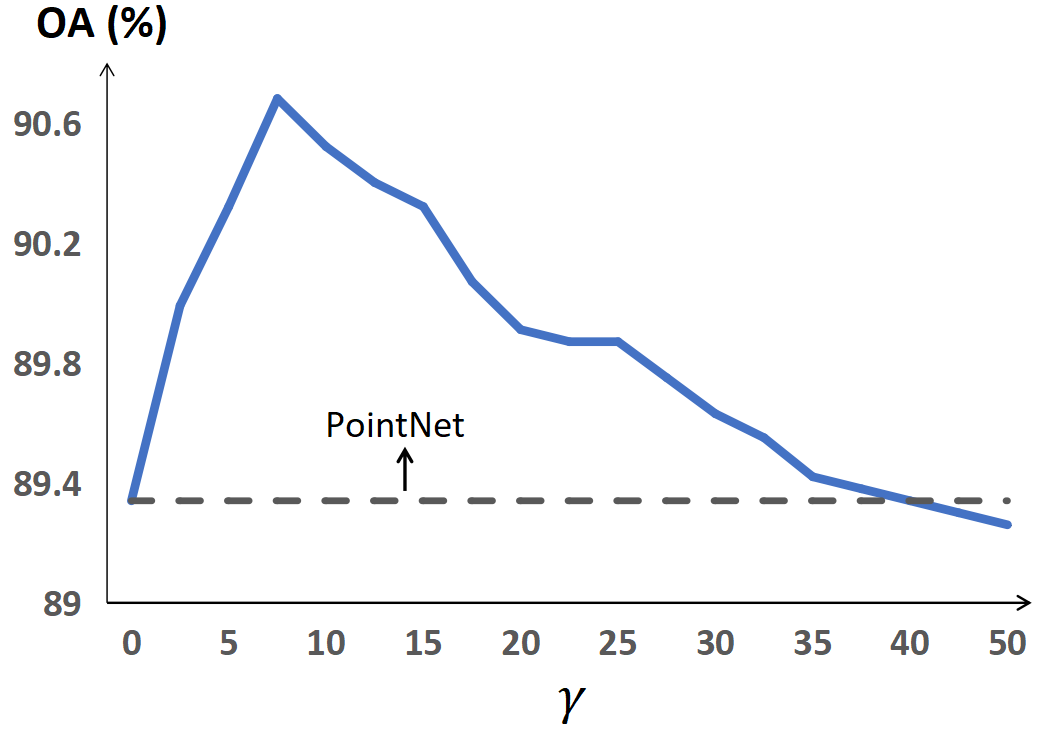}
    \caption{Ablation study of \textbf{interpolation ratio $\gamma$}.}
    \label{figure4}
\end{figure}

\begin{figure}[t!]
    \centering
    \includegraphics[height=1.8in]{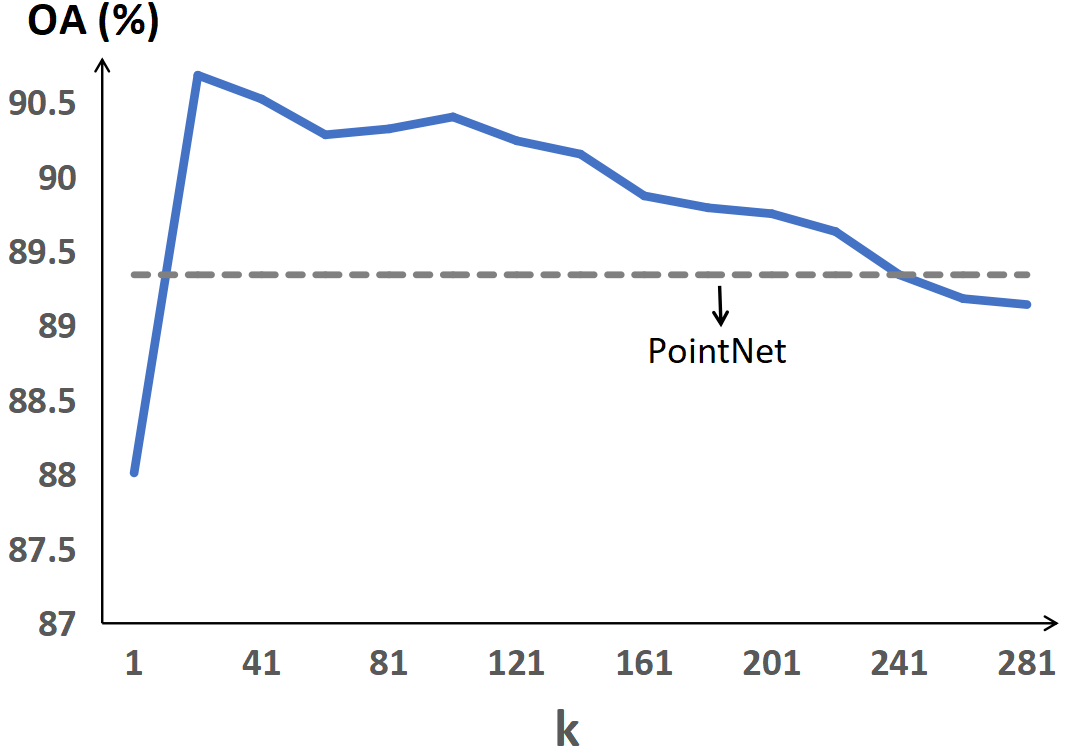}
    \caption{Ablation study of \textbf{the number of nearest neighbors, $k$}.}
    \vspace{0.15cm}
    \label{figure5}
\end{figure}

\paragraph{Main hyperparameters}
We here conduct the ablation study concerning two hyperparameters: $\gamma$ and $k$. We adopt PointNet~\cite{qi2017pointnet} with SN-Adapter and experiment shape classification on ModelNet40~\cite{modelnet}. As the $\gamma$ varies from 0 to 50 in Figure~\ref{figure4}, the enhancement of SN-Adapter peaks around 8, but becomes harmful after 10. This indicates the SN-Adapter would adversely affect the baseline model under a too large proportion, and requires a proper interpolation ratio for best introducing the spatial prototypical knowledge. The results in Figure~\ref{figure5} show that our SN-Adapter is not very sensitive to $k$ when it is large enough (over 80), which has already covered the most contributing prototypes to the final classification.

\begin{table}[t]
\vspace{0.2cm}
\small
\centering
	\begin{tabular}{lccc}
	\toprule
		\makecell*[c]{Metric} &PointNet &DGCNN &CurveNet\\
		\cmidrule(lr){1-1} \cmidrule(lr){2-2} \cmidrule(lr){3-3} \cmidrule(lr){4-4}
	    Manhattan & 90.36 & 92.63 & 94.00 \\
 Chebyshev & 88.29 & 92.26 & 93.56 \\
 Hamming & 88.37 & 92.22 & 93.72 \\
 Canberra & 90.07 & 91.82 & 93.92 \\
 Braycurtis & 90.24 & 92.46 & 94.04 \\
 Euclidean & \textbf{90.68} & \textbf{92.99} & \textbf{94.25} \\
	  \bottomrule
	\end{tabular}
\vspace{0.4cm}
\caption{\textbf{Different distance metrics} for SN-Adapter on ModelNet40~\cite{modelnet40} dataset with overall accuracy (OA) (\%).}
\label{t6}
\end{table}

\paragraph{Distance metrics for retrieval.}
Different distance metrics for SN-Adapter affect the retrieval of nearest spatial prototypes, which further leads to different performance enhancement over the baseline models. We evaluate our SN-Adapter with different distance metrics for shape classification on ModelNet40~\cite{modelnet} and adopt three baseline models: PointNet~\cite{qi2017pointnet}, DGCNN~\cite{dgcnn}, and CurveNet~\cite{curvenet}. As reported in Table~\ref{t6}, for three baseline models, Euclidean distance performs better, which can be better reveal the point distribution in the 3D space.

\paragraph{Positional encodings.}
For shape classification, we equip the sample-wise $Proto^{cls}$ with global positional vectors to preserve the spatial distributions of points. In Table\ref{t7}, we explore the best way to obtain such vectors concerning the encoding functions and pooling operations. We evaluate three baseline models: PointNet~\cite{qi2017pointnet}, PointNet++~\cite{qi2017pointnet++} and PCT~\cite{guo2021pct} for shape classification on ModelNet40~\cite{modelnet}. As reported, 'Sin/cos' encoding function has more advantages, which can bring favorable performance boost to the `SN-Adapter without PE'.

\begin{table}[t!]
\small
\centering
\vspace{0.1cm}
	\begin{tabular}{ccccc}
	\toprule
		\makecell*[c]{PE} &Pooling &PointNet &PointNet++ &PCT\\
		\cmidrule(lr){1-1} \cmidrule(lr){2-2} \cmidrule(lr){3-3} \cmidrule(lr){4-4} \cmidrule(lr){5-5}
 - &- &90.11 & 93.19 & 93.52 \\
 Fourier &Avg. & 89.99 & 93.11 & 93.52 \\
 Fourier &Max. & 89.95 & 93.23 & 93.48 \\
 Sin/cos &Avg. & 90.03 & \textbf{93.48} & \textbf{93.56} \\
 Sin/cos &Max. & \textbf{90.68} & 93.15 & 93.48\\
	  \bottomrule
	\end{tabular}
\vspace{0.4cm}
\caption{\textbf{Different positional encodings (PE) and pooling operations} of SN-Adapter for on ModelNet40~\cite{modelnet40} dataset with overall accuracy (OA) (\%). `Fourier' and `Sin/cos' denote Fourier and trigonometric encoding functions~\cite{3detr}, respectively. The first row denotes SN-Adapter without any positional encodings.}
\label{t7}
\end{table}

\begin{figure*}[t!]
  \centering
    \includegraphics[width=0.95\textwidth]{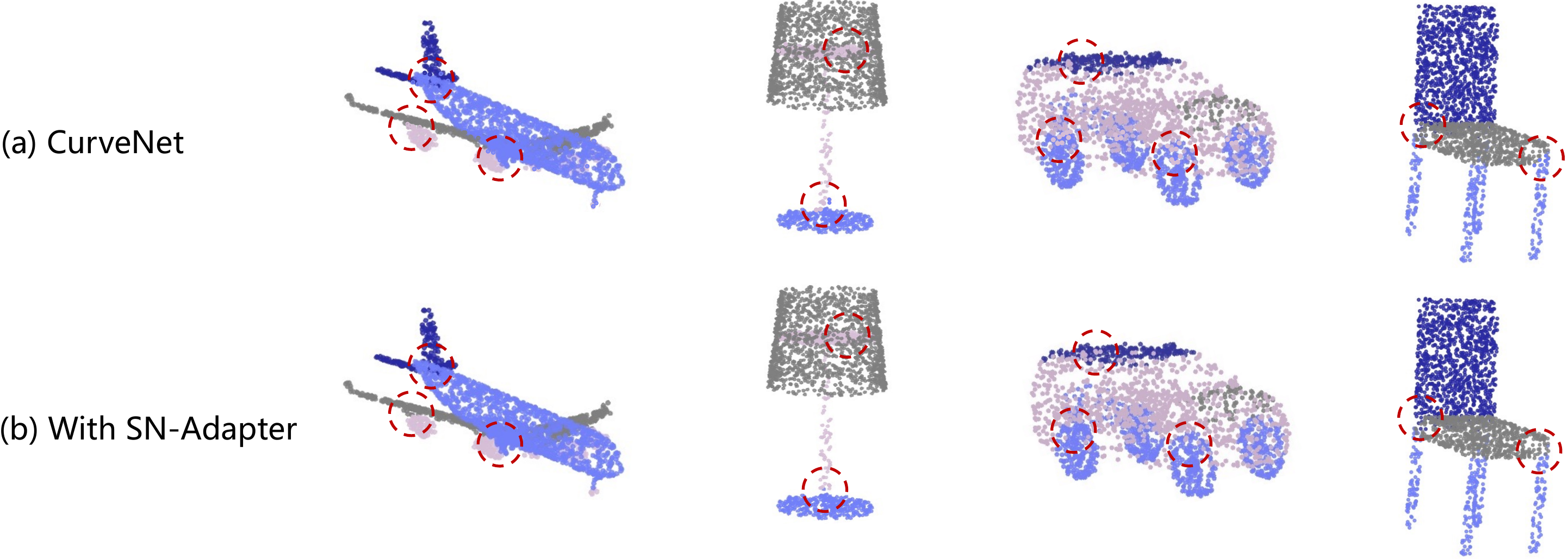}
    \vspace{0.3cm}
   \caption{\textbf{Visualization of part segmentation without (a) and with (b) our SN-Adapter} on ShapeNetpart~\cite{shapenetpart} dataset. We select CurveNet~\cite{curvenet} as the baseline model and highlight the differences by red circles (Zoom in for a better view).}
  \vspace{0.1cm}
    \label{figv}
\end{figure*}

\begin{table}[t!]
\small
\centering
\vspace{0.1cm}
	\begin{tabular}{lcc}
	\toprule
		\makecell*[c]{Method} &AP$_{25}$ (\%) &AR$_{25}$ (\%)\\
		\cmidrule(lr){1-1} \cmidrule(lr){2-2} \cmidrule(lr){3-3}
		3DETR-m  &64.60 &77.22\\
	    3DETR-m + SN-Adapter  &\textbf{65.16} &\textbf{84.56}\\
	    After 3D NMS &64.62 &78.19\\
	    Without PE &65.02 &83.48\\
	  \bottomrule
	\end{tabular}
\vspace{0.2cm}
\caption{Ablation study of \textbf{SN-Adapter for 3D object detection} on ScanNetV2~\cite{ScanNetV2}. For last two rows, we respectively insert SN-Adapter after 3D NMS and discard the positional encodings.}
\label{t8}
\end{table}

\paragraph{3D Object Detection}
We insert our SN-Adapter into the trained object detectors before 3D NMS, and summarize the object-wise prototypes with positional encodings. We here explore the effectiveness of both insert position and positional encodings. In Table~\ref{t8}, we select 3DETR-m~\cite{3detr} as our baseline on ScanNetV2~\cite{ScanNetV2} dataset. As shown, if after 3D NMS, the SN-Adapter cannot bring noteworthy boosts, since the remaining 3D boxes filtered by NMS are already the most confident ones for the detector. Also, blending the positional encodings can improve the performance of SN-Adapter for introducing more positional knowledge into the prototypes.

\paragraph{Extra Costs of SN-Adapter.}
Besides the enhancement on scores, we explore if our SN-Adapter would cause too much extra time and memory costs over baseline models. We utilize a single RTX 3090 GPU with batch size 64 for testing and select two baseline models: DGCNN~\cite{dgcnn} for shape classification on ModelNet40~\cite{modelnet40} and CurveNet~\cite{curvenet} for part segmentation on ShapeNetPart~\cite{shapenetpart}. As shown in Table~\ref{t9}, our non-parametric SN-Adapter can achieve superior performance-cost trade-off to enhance already trained networks without re-training.

\subsection{Visualization}
In Figure~\ref{figv}, we visualize the results of CurveNet~\cite{curvenet} with and without our SN-Adapter for part segmentation on ShapeNetPart~\cite{shapenetpart} dataset. As shown, our SN-Adapter mainly improves the segmentation of points located in the connection areas between different object parts. Such points normally contain the semantic knowledge of both object parts and would confuse the learned 3D classifier of deep neural networks. In contrast, our SN-Adapter could alleviate such issue by retrieving from the prototypes to obtain better part-wise discrimination capability.

\begin{table}[t]
\small
\centering
	\begin{tabular}{lccc}
	\toprule
		\makecell*[c]{Method} &Score (\%)  &Latency &Memory\\
		\cmidrule(lr){1-1} \cmidrule(lr){2-2} \cmidrule(lr){3-3} \cmidrule(lr){4-4}
	    DGCNN & 92.18 & 0.022s & 9.74 GiB \\\rowcolor{gray!12}\vspace{0.1cm}
	    \ \ \ \ + SN-Adapter &\textbf{92.99} & 0.046s & 10.06 GiB \\
	    CurveNet & 86.58 & 0.607s & 10.93 GiB \\\rowcolor{gray!12}\vspace{0.1cm}
	    \ \ \ \ + SN-Adapter &\textbf{86.69} & 0.834s &11.50 GiB \\
	  \bottomrule
	\end{tabular}
\vspace{0.2cm}
\caption{\textbf{The extra costs of SN-Adapter for time and memory.} We test on a single RTX 3090 GPU with batch size 64 and report the OA/ mIoU$_I$ for DGCNN/ CurveNet.}
\label{t9}
\end{table}

\section{Conclusion}
We propose Spatial-Neighbor Adapter (SN-Adapter), a plug-and-play enhancement module for existing 3D networks without extra parameters and re-training. From the pre-constructed prototypes, SN-Adapter leverages $k$ nearest neighbors to retrieve spatial knowledge and effectively boost the 3D networks by providing complementary characteristics. \textbf{Limitations.} Although our SN-Adapter can be generalized to various tasks, \eg, shape classification, part segmentation, and 3D object detection, the performance enhancement for part segmentation is relatively lower than others. Our future work will focus on designing more advanced part-wise prototypes for better segmentation results.

{\small
\bibliographystyle{ieee_fullname}
\bibliography{egbib}
}

\end{document}